# Accelerated and Inexpensive Machine Learning for Manufacturing Processes with Incomplete Mechanistic Knowledge

Jeremy Cleeman, Kian Agrawala, Rajiv Malhotra*

Department of Mechanical and Aerospace Engineering, Rutgers University, 98 Brett Road, Piscataway, NJ, USA

*Corresponding author: rajiv.malhotra@rutgers.edu

**Abstract:** Machine Learning (ML) is of increasing interest for modeling parametric effects in manufacturing processes. But this approach is limited to established processes for which a deep physics-based understanding has been developed over time, since state-of-the-art approaches focus on reducing the experimental and/or computational costs of generating the training data but ignore the inherent and significant cost of developing qualitatively accurate physics-based models for new processes . This paper proposes a transfer learning based approach to address this issue, in which a ML model is trained on a large amount of computationally inexpensive data from a physics-based process model (source) and then fine-tuned on a smaller amount of costly experimental data (target). The novelty lies in pushing the boundaries of the qualitative accuracy demanded of the source model, which is assumed to be high in the literature, and is the root of the high model development cost. Our approach is evaluated for modeling the printed line width in Fused Filament Fabrication. Despite extreme functional and quantitative inaccuracies in the source our approach reduces the model development cost by years, experimental cost by 56-76%, computational cost by orders of magnitude, and prediction error by 16-24%.

## 1. Introduction

Machine Learning (ML) models have become popular for modeling parametric effects in manufacturing processes due to their high deployability. But generating the required training data from experiments incurs time and resource expenditure (experimental cost $C_E$). Generating the training data from physics-based process models incurs a computational cost $C_C$, i.e., CPU-hours needed to run simulations; and a model development cost $C_D$, i.e., the time and human resources needed for intuitive trial-and-error creation of constitutive laws and numerical methods that qualitatively and quantitatively capture interactions between multiple physical phenomena over multiple time and length scales.[1] The root cause of high $C_D$ for new processes, which is often on the order of decades[2-5], is that qualitative knowledge of the underlying physics is often missing.

Multifidelity learning trains an ML model using a large amount of inexpensive and inaccurate data (source), and fine-tunes it using a small amount of costly but accurate data (target). Using computational process models as the source and experimental data as the target reduces $C_E$ relative to training with only experimental data, reduces $C_C$ compared to training with only computational data, and captures the ground truth.[6] But these works assume that the source must qualitatively match the target, i.e., multifidelity learning only effects a quantitative correction. Thus, $C_D$ is still high since a qualitatively accurate physics-based source is needed. Using analytical process models as the source decreases $C_C$ even further, but does not reduce $C_D$.[7] Note that using experimental sources for new processes is not possible due to their inherent novelty.

This paper proposes a multifidelity learning approach that reduces $C_D$ despite limited mechanistic knowledge of the process physics. This method is demonstrated for modeling the printed line's width *W* in Fused Filament Fabrication (FFF) as a function of the filament feed rate *F* and extruder speed *S*. This problem involves complex physics including non-Newtonian flow, friction, cooling, wetting, and compressibility.[8,9] While FFF is not new and this problem remains solved today, we choose this problem since it is this very fact that allows us to quantify the reduction in $C_D$ possible.

## 2. Methods

Our approach use transfer-based multifidelity learning, with a physics-based process model as the source and experiments as the target. Thus, the final ML model reflects the experimental ground truth and reduces $C_E$. This process model must (a) include one or more conservation laws to respect the fundamental laws of nature; (b) use a guess for the form of the constitutive law without any experimental calibration or validation, to reduce $C_D$; (c) avoid or minimize spatiotemporal discretization to minimize $C_C$. The reader is referred to the literature for further details on the various transfer learning methods available for regression.[10,11] In this paper, Epsilon Support Vector Regression (SVR[12,13]) with a gaussian Radial Basis Function was used as the ML model, and TrAdaBoostR2 instance-based transfer learning was used for fine-tuning.[14] The hyperparameters for the ε-SVR were based on brute force identification and the number of boosting iterations for TrAdaBoostR2 was 30. The reader is referred to the above literature for further details on both SVRs and TrAdaBoostR2.

The source model was the mass conservation law, i.e., $W = FA/Sh$, where *h* is the nozzle-to-platen distance and *A* is the filament's cross-sectional area. This model ignores the complexity of almost all the earlier mentioned extrusion physics, and makes an incorrect but simplifying assumption that *h* equals the line (or layer) height. It took ≈ $10^{-6}$ CPU-hours to generate the 624 source samples used here. Experiments were performed to print PLA lines on a home-built FFF machine with a 1 mm diameter nozzle for sixteen equidistant *S* (between 350 and 725 mm/min) and *F* (between 153 and 729 mm/min) across *h* = 0.7, 0.85, 1.2 mm. The *W* was measured using vernier calipers and averaged across 3 measurements. Unstable printing regimes were excluded.

First, direct learning of the SVR was performed on only the experimental target data. Progressively more training points were used till the Root Mean Square Error (RMSE) on the testing data (i.e., the remainder of the dataset) did not decrease further. This training and testing was performed 1000 times using random sampling, and yielded the average values of the smallest error $RMSE_{direct}$ and the corresponding number of samples $n_{direct}$ for direct learning. Transfer learning was performed with the source data of the same size as $n_{direct}$. A progressively increasing amount of target data was used to iteratively identify the smallest target dataset needed for transfer learning ($n_t$) such that the transfer learning error $RMSE_t$ was lesser than or equal to $RMSE_{direct}$. This ensured that prediction accuracy was not sacrificed in the drive to reduce costs. Testing of the final SVR obtained after transfer learning was performed on data obtained randomly from the portion of the experimental data not used for training. This test dataset was of the same size as $n_{direct}$ in order to prevent a heavily lopsided train:test ratio and thus fairly compare direct l and transfer learning. This randomized testing was performed 30 times to obtain the mean $RMSE_t$.

## 3. Results

Figure 1 shows the functional discrepancy between the source model and the experimental target with 3D plots and representative 2D plots. The true effect of *F* and *S* on *W* is decidedly nonlinear, especially at lower *h*, as compared to the linear assumption in the source. Figures 2a-c show the change in the tested RMSE of direct learning on only experimental data as a function of the number of training points, and reveals the $RMSE_{direct}$ and $n_{direct}$ (which is constant at 150 for all $h$). Figures 2d-f compare this $RMSE_{direct}$ to the error from transfer learning for different amounts of experimental data (i.e., combinations of $F$ and $S$). There are multiple cases for which transfer learning enables $RMSE_t \leq RMSE_{direct}$ and $n_t < n_{direct}$. Qualitatively, Figure 3 shows that the transfer learnt SVR can capture the nonlinearity in the experimental data despite the qualitatively and quantitatively inaccurate mechanistic knowledge embedded in the source model.

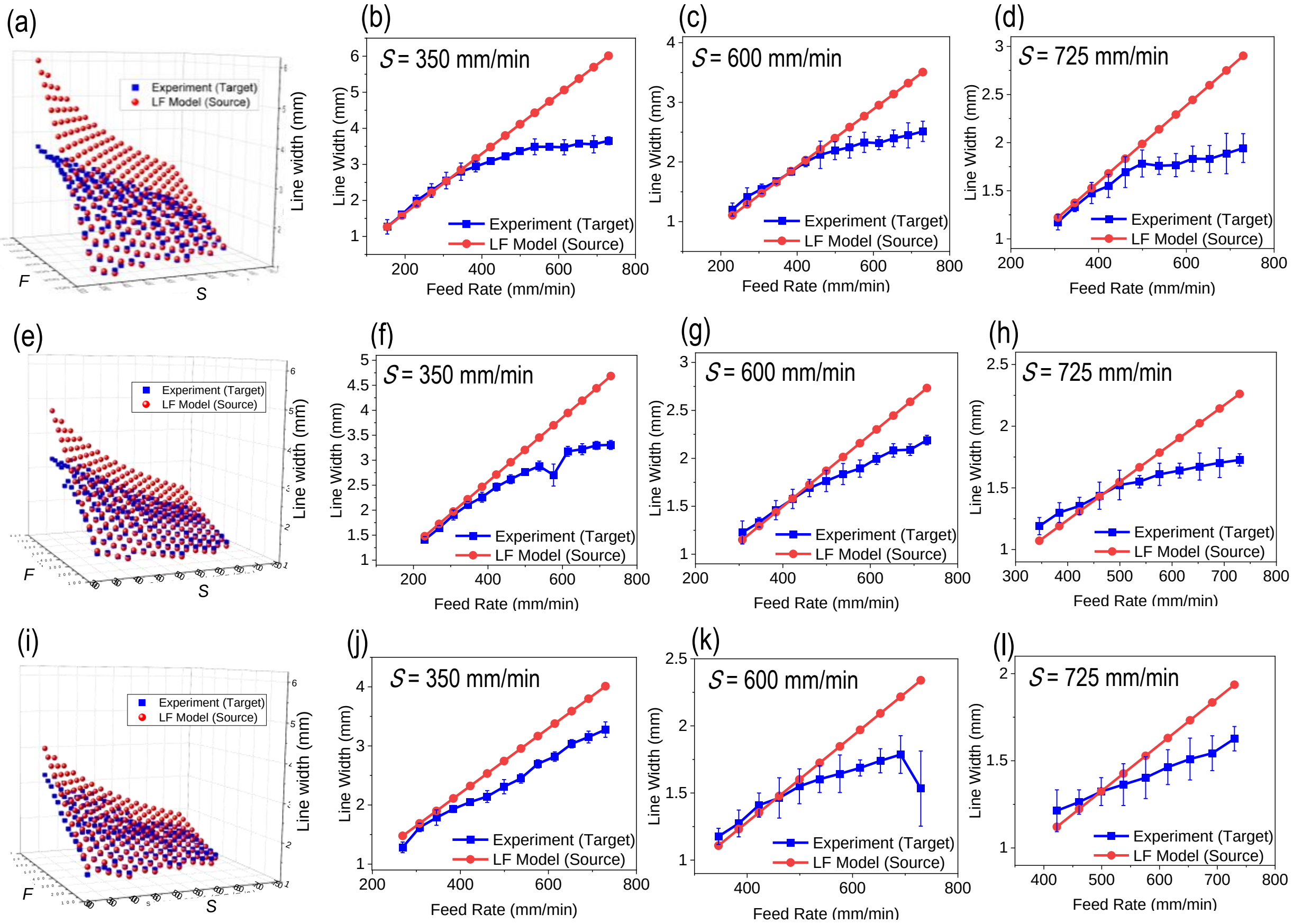

**Figure 1:** Comparison of source and target for $h$ = (a) 0.7 mm (b) 0.85 mm (c) 1.2 mm. Feed rate $F$ and stage speed $S$ are in mm/min.

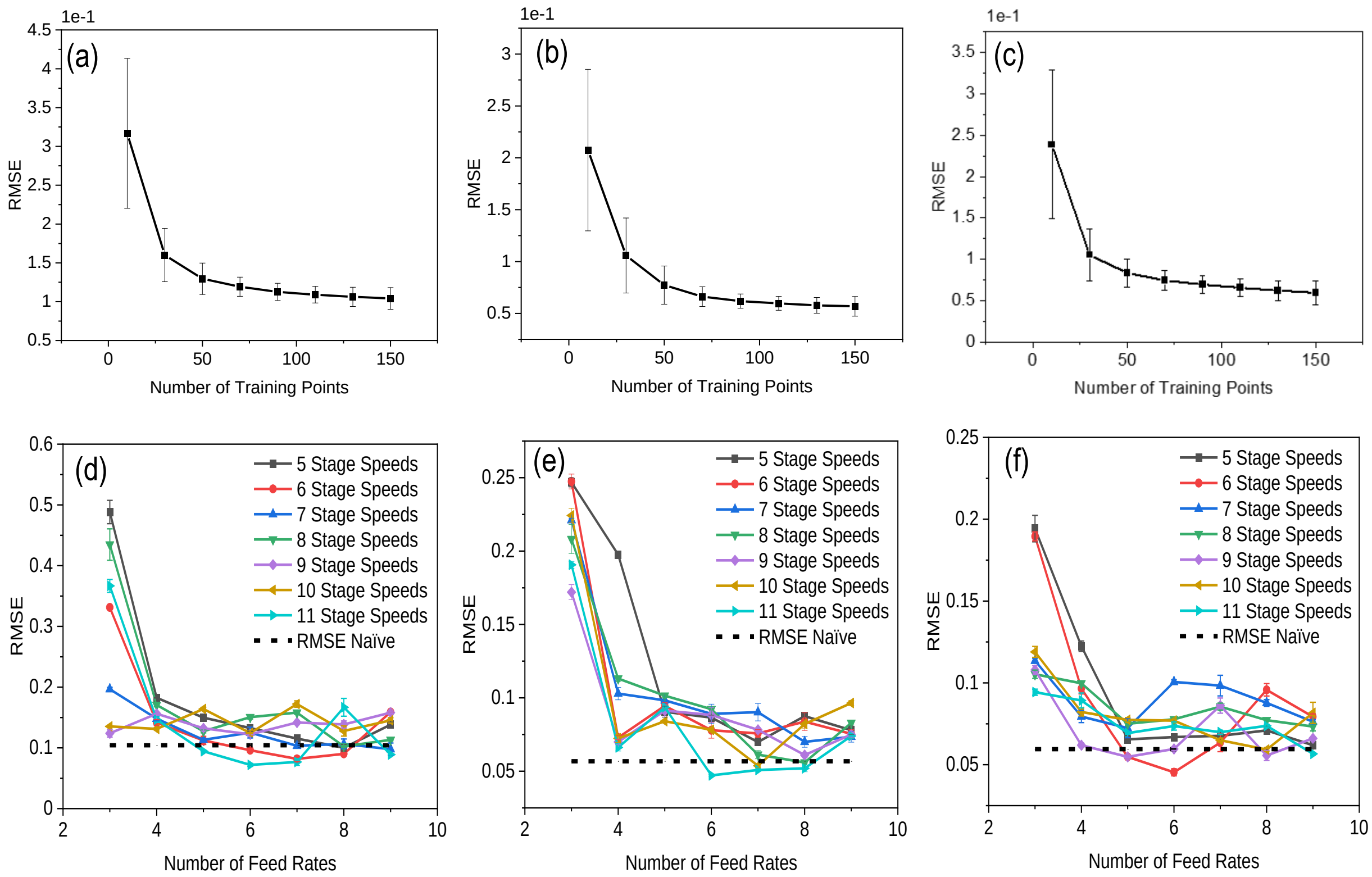

**Figure 2:** RMSE from direct learning as a function of the number of training points for for *h* = (a) 0.7 mm (b) 0.85 mm (c) 1.2 mm. Comparison of $RMSE_{direct}$ to the error obtained from transfer learning using different amounts of experimental *F* and *S* and for *h* = (d) 0.7 mm (e) 0.85 mm (f) 1.2 mm. Feed rate *F* and stage speed *S* are in mm/min.

Our approach realizes a 56-76 % reduction in $C_{EXP}$ as compared to direct learning on experimental data, and reduces the error by 16-24% (Table 1). Already developed computational or analytical process models can be used as the source or for direct learning, since they are good qualitative and quantitative matches to the ground truth.[8,9] But it has taken significant time and effort for these models to reach this point, from 2000-2019 for analytical equations[15,16] and from 2002 to 2018 for computational simulations[9,17]. This indicates that using Smart-ML in 2000, which is when our source model was reported in the literature, could have saved at least 15 human-years of $C_{DEV}$. Overall, our approach reduces $C_{DEV}$ for new processes by easing the need for qualitatively accurate human-created physics-based process models. Note that using high-fidelity computational models to generate just one training sample for FFF needs orders of magnitude more CPU-hours than that for Smart-ML (i.e., $10^{-6}$ CPU-hours).[9,16] Thus, Smart-ML reduces $C_{DEV}$ in addition to $C_{COMP}$ and $C_{EXP}$.

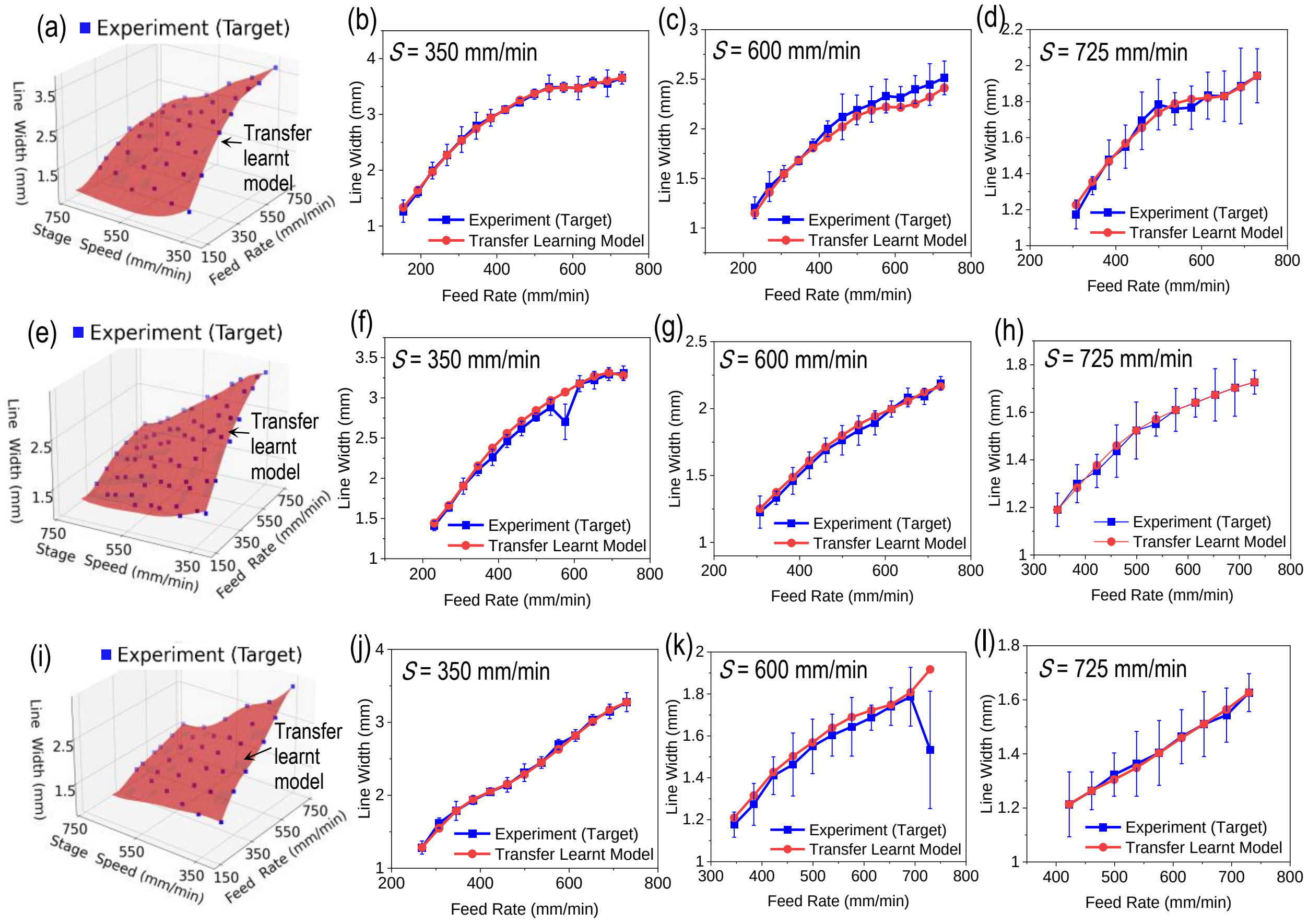

**Figure 3:** Comparison of transfer learnt model and target for $h$ = (a-d) 0.7 mm (e-h) 0.85 mm (i-l) 1.2 mm.

Table 1. Comparison of smallest RMSE and corresponding number of training samples for direct learning and transfer learning

| $h$ (mm) | $n_{direct}$ | $RMSE_{direct}$ | $n_t$ | | $RMSE_t$ | $\frac{n_{naive} - n_t}{n_{naive}}$ | $\frac{RMSE_{naive} - RMSE_t}{RMSE_{naive}}$ |
|---|---|---|---|---|---|---|---|
| | | | No. of $S$ | No. of $F$ | | | |
| 0.7 | 150 | 0.104± 0.014 | 6 | 7 | 0.081± 0.004 | 72% | 22% |
| 0.85 | 150 | 0.056± 0.009 | 11 | 6 | 0.047± 0.0006 | 56% | 16% |
| 1.2 | 150 | 0.059 ± 0.015 | 6 | 6 | 0.045± 0.002 | 76% | 24% |

## 4. Conclusions

State-of-the-art approaches for ML models of parametric effects in manufacturing processes focus on reducing the experimental and computational cost of training data generation. This paper pushes beyond this paradigm to examine the possibility of also reducing the often-overlooked, but significant, cost of process model development. This is achieved by testing the limits of the requisite similarity between source process models and target experimental data in transfer learning, by exploring the use of an uncalibrated guess for the functional form of the constitutive law to avoid the cost of iterative model development. This approach overcomes significant functional discrepancies between the source and the target, unlike assumptions made in the manufacturing literature; reduces the developmental cost along with the experimental, and computational costs of generating training data; and reduces the prediction error.